\documentclass[11pt,a4paper]{article}
\usepackage{times,latexsym}
\usepackage{url}
\usepackage[T1]{fontenc}

\usepackage[acceptedWithA]{tacl2021v1}

\usepackage{amsmath}
\usepackage{booktabs}
\usepackage{siunitx}
\usepackage{mathtools}
\usepackage[ruled,vlined,noend]{algorithm2e}
\usepackage{graphicx}
\usepackage{mathtools}
\usepackage{bbm}
\usepackage{amsfonts}
\usepackage[inline]{enumitem}
\usepackage{tabularx}
\usepackage{arydshln}
\usepackage{multirow}
\usepackage{makecell}
\usepackage{xcolor}
\usepackage{soul}
\usepackage[inline]{enumitem}
\usepackage{adjustbox}
\usepackage{csquotes}

\SetKwIF{If}{ElseIf}{Else}{if}{}{else if}{else}{end if}%
\SetKwFor{While}{while}{}{end while}%
\SetKwRepeat{Do}{do}{while}
\SetKw{KwGoTo}{go to}
\usepackage{mathtools}

\newif\ifcomments
\commentstrue
\ifcomments
    \providecommand{\mi}[1]{{\protect\color{olive}{[MI: #1]}}}
    
    \def \ifempty#1{\def\temp{#1} \ifx\temp\empty }

    \providecommand{\jb}[1]{{\protect\color{blue}{[JB: #1]}}}
    \providecommand{\yair}[1]{{\protect\color{magenta}{[YC: #1]}}}
    \providecommand{\ol}[1]{{\protect\color{cyan}{[OL: #1]}}}
    \providecommand{\mg}[1]{{\protect\color{orange}{[MG: #1]}}}
    \providecommand{\remove}[1]{{\protect\color{purple}{[RM: #1]}}}
    \providecommand{\us}[1]{{\protect\color{violet}{[US: #1]}}}

\else
    \providecommand{\mi}[1]{}
    \providecommand{\jb}[1]{}
    \providecommand{\yair}[1]{}
    \providecommand{\ol}[1]{}
    \providecommand{\mg}[1]{}
    \providecommand{\remove}[1]{}
    \providecommand{\us}[1]{}
\fi
\providecommand{\commentout}[1]{}

\newcommand\SCROLLS{\textsc{SCROLLS}}
\newcommand\localhypo{\emph{Locality of information assumption}}

\newcommand\bartbase{\textsc{BART}$_\mathtt{base}$}
\newcommand\bartlarge{\textsc{BART}$_\mathtt{large}$}

\newcommand\sled{\textsc{SLED}}

\newcommand\blsled{\bartlarge{}-\sled{}}
\newcommand\led{\textsc{LED}}
\newcommand\ledbase{\textsc{LED}$_\mathtt{base}$}
\newcommand\ledlocal{\textsc{LED}$^\mathcal{L}$}

\newcommand\ledbaseglobal{\textsc{LED}$_\mathtt{base}^\mathtt{SCROLLS}$}



\title{Efficient Long-Text Understanding with Short-Text Models}

\author{Maor Ivgi ~~~~~~Uri Shaham ~~~~~~
Jonathan Berant \\
The Blavatnik School of Computer Science, Tel-Aviv University \\
\small{\texttt{\{maor.ivgi,uri.shaham,joberant\}@cs.tau.ac.il}}}

\date{}

\begin{document}
 \maketitle

\begin{abstract}

Transformer-based pretrained language models (LMs) are ubiquitous across natural language understanding, but cannot be applied to long sequences such as stories, scientific articles and long documents, due to their quadratic complexity. While a myriad of efficient transformer variants have been proposed, they are typically based on custom implementations that  require expensive pretraining from scratch.
In this work, we propose \textbf{\sled{}}: \textbf{SL}iding-\textbf{E}ncoder and \textbf{D}ecoder, a simple approach for processing long sequences that re-uses and leverages 
battle-tested short-text pretrained LMs. Specifically, we partition the input into overlapping chunks, encode each with a short-text LM encoder and use the pretrained decoder to fuse information across chunks (\emph{fusion-in-decoder}).
We illustrate through controlled experiments that \sled{} offers a viable strategy for long text understanding and evaluate our approach on \SCROLLS{}, a benchmark with seven datasets across a wide range of language understanding tasks.
We find that \sled{} is competitive with specialized models that are up to 50x larger and require a dedicated and expensive pretraining step.

\end{abstract}
\section{Introduction}
\label{sec:intro}

\begin{figure}[t]
\begin{center}

\centerline{
\includegraphics[width=\linewidth]{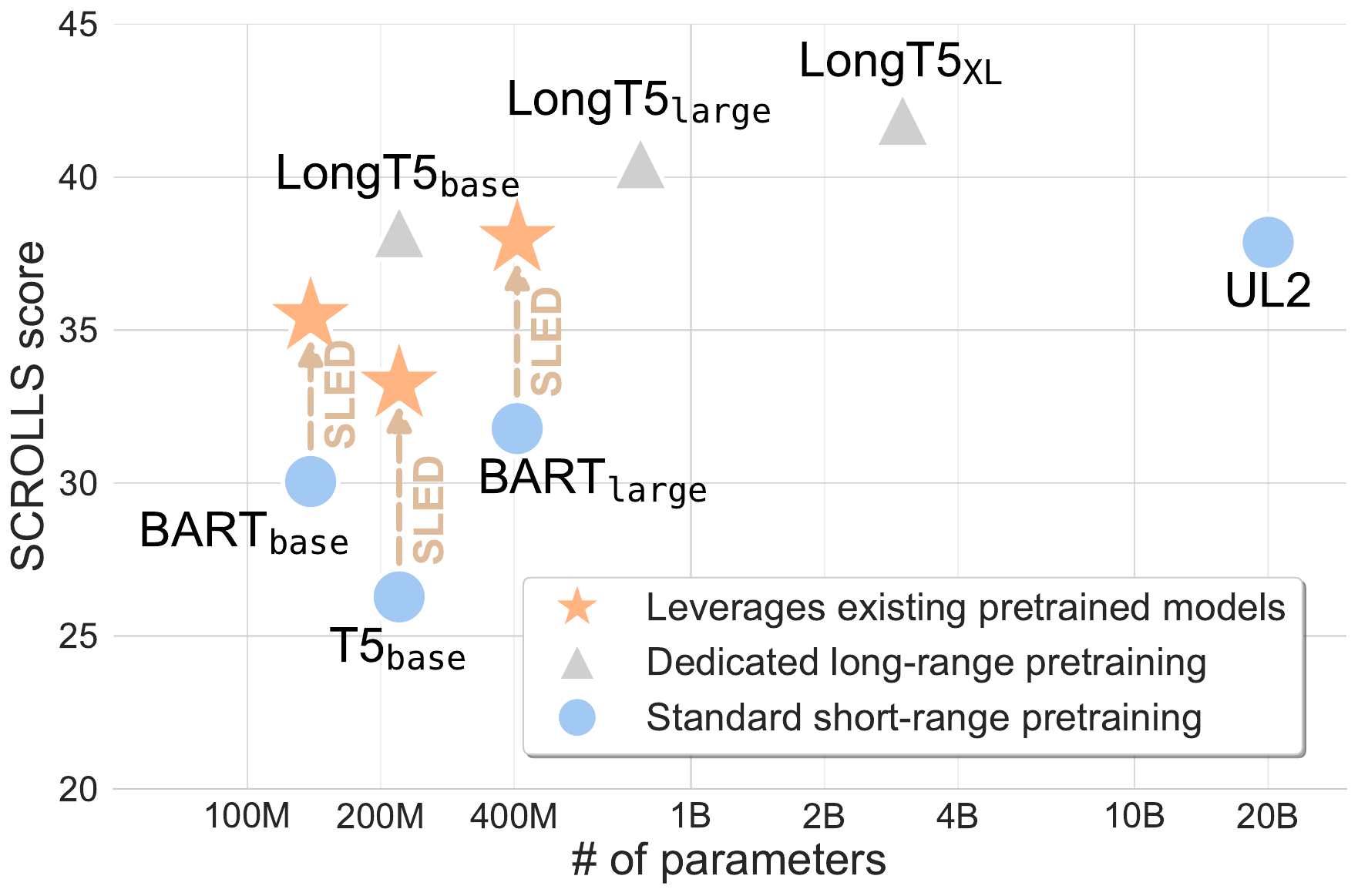}
}
\setlength{\belowcaptionskip}{-20pt}
\caption{Models' \SCROLLS{} score
\cite{Shaham2022SCROLLSSC} as a function of parameter count. Plugging existing pretrained LMs into the \sled{} framework dramatically improves their \SCROLLS{} score (arrows from blue circles to orange stars).
Gray triangles indicate models with dedicated pretraining for capturing long-range dependencies. BART$_\text{large}$-\sled{} is competitive with LongT5$_\text{base}$ \cite{Guo2021LongT5ET} and UL2 \cite{Tay2022UnifyingLL} (which has 50x more parameters), and slightly lags behind larger LongT5 models.
}

\label{fig:scrolls-score}
\end{center}
\end{figure}
Transformer-based pretrained language models
\cite{Vaswani2017AttentionIA,Devlin2019BERTPO,Lewis2020BARTDS,Raffel2020ExploringTL,Brown2020LanguageMA} 
have been widely successful across all areas of natural language understanding (NLU).
However, applying them over long texts (such as stories, scripts, or scientific articles) is prohibitive due to their quadratic complexity in the input length.
To bridge this gap, recent work has developed more efficient transformer variants \cite{Kitaev2020ReformerTE,beltagy2020longformer,Zaheer2020BigBT,Guo2021LongT5ET} and applied them over long-range language understanding tasks \cite{Mehta2022LongRL,Shaham2022SCROLLSSC}.

However, most efficient transformers use specialized architectures with custom implementations that are not guaranteed to scale as well as vanilla transformers \cite{Tay2022ScalingLV}. Moreover, they require an expensive pretraining step and do not exploit off-the-shelf pretrained LMs that were trained for short texts. To date, their performance on long texts has not matched the success of their short-range counterparts.

\begin{figure*}[t]
\begin{center}

\centerline{
\includegraphics[width=\linewidth,trim={0.5cm 0.5cm 11cm 22.1cm},clip]{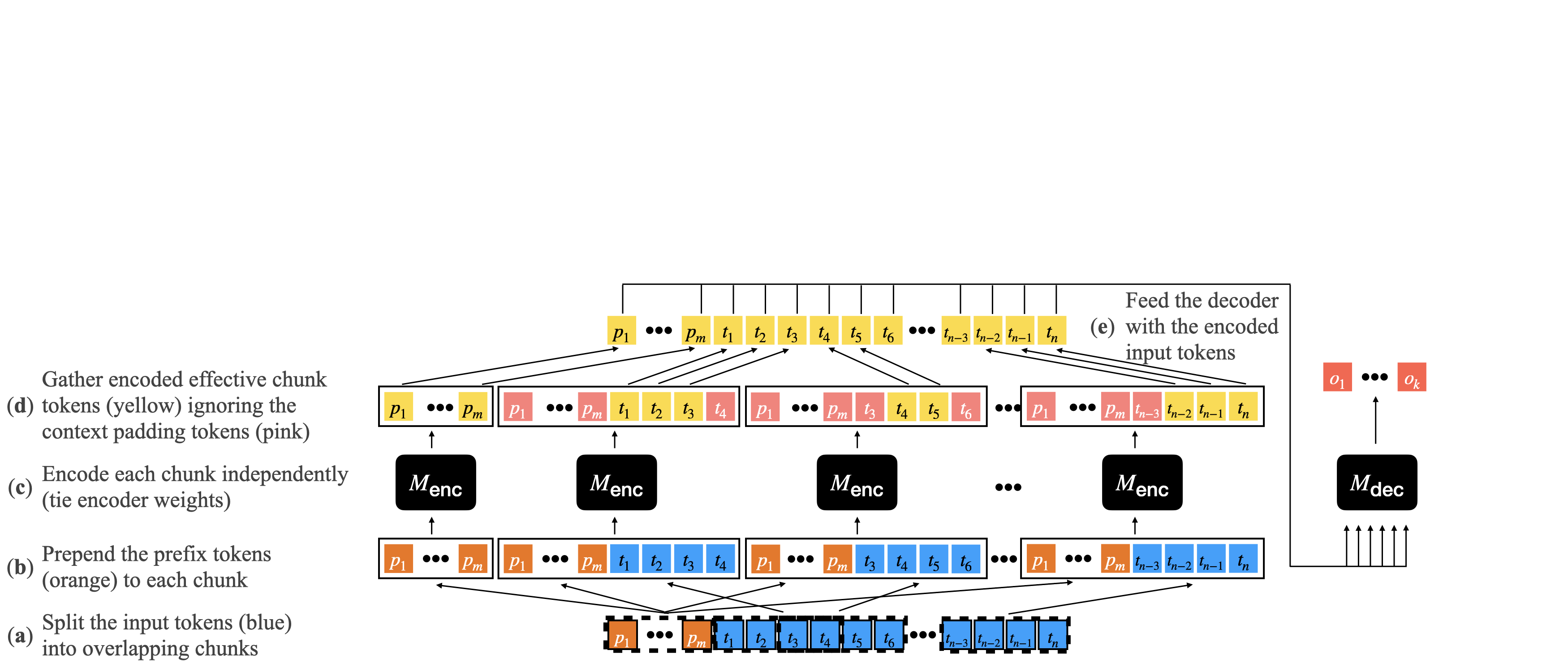}}

\setlength{\belowcaptionskip}{-20pt}
\caption{Overview of \sled{}. (a) Input tokens $(t_1,\dots,t_n)$ are chunked into $C$ overlapping chunks of length $c$ (here, $c=4$). Each chunk is made of $P:=\frac{\rho \times c}{2}$ \emph{context padding} tokens at the right and left edges of the chunk, and $(1-\rho) \times c$ \emph{effective chunk} tokens in the middle (here, $\rho=0.5, P=1$). (b) We prepend the prefix tokens $(p_1,\dots,p_m)$ to each chunk ($m \ll n$). (c) Each chunk is encoded independently using the already pretrained backbone encoder $M_{\text{enc}}$. (d) We gather the encoded \emph{effective chunks} tokens (yellow) and discard the \emph{context padding} tokens (pink) (e) We pass the encoded input to the decoder to generate the final output sequence $(o_1,\dots,o_k)$.
}

\label{fig:sled}
\end{center}
\end{figure*}

In this work, we present \textbf{\sled{}}: \textbf{SL}iding-\textbf{E}ncoder and \textbf{D}ecoder, a simple yet powerful method for applying off-the-shelf pretrained encoder-decoder models on long text problems, with a linear time and space dependency. Specifically (see Fig.~\ref{fig:sled}), we
partition long documents into overlapping chunks of tokens of constant length and encode each chunk independently with an already-pretrained encoder. Then, a pretrained decoder attends to all contextualized input representations to generate the output. Our main assumption is that input tokens can be contextualized through their local surrounding (using a short-text LM), and any global cross-chunk reasoning can be handled by the decoder, similar to fusion-in-decoder (FiD) \cite{Izacard2021LeveragingPR}.
Our approach can be readily applied to \emph{any} pretrained encoder-decoder LM such as T5 \cite{Raffel2020ExploringTL} and \textsc{BART} \cite{Lewis2020BARTDS} (but is not applicable to decoder-only \cite{Brown2020LanguageMA} or encoder-only models \cite{Liu2019RoBERTaAR,Conneau2020UnsupervisedCR}).

We evaluate \sled{} on a wide range of language understanding tasks. To substantiate \sled{}'s adequacy for text processing, we perform controlled experiments over modified versions of
SQuAD 1.1 \cite{Rajpurkar2016SQuAD1Q} and HotpotQA \cite{Yang2018HotpotQAAD} to show that \sled{} can (a) find relevant information that is embedded within a long text sequence and (b) fuse information from chunks that were encoded separately. 

Our main evaluation is over \SCROLLS{}, a recently-released benchmark that includes 7 long-range tasks across Question Answering (QA), Summarization, and Natural Language Inference (NLI). We show (Fig.~\ref{fig:scrolls-score}) that taking a pre-trained encoder-decoder model, such as BART \cite{Lewis2020BARTDS} or T5 \cite{Raffel2020ExploringTL}, and embedding it into \sled{}'s framework results in dramatic improvement in performance (6 points on average across models). Moreover, \blsled{}'s performance is comparable to LongT5$_{\mathtt{base}}$ \cite{Guo2021LongT5ET}, a model that was specifically pretrained to handle long-range dependencies, and surpasses UL2 \cite{Tay2022UnifyingLL}, which contains 50x more parameters.
Importantly, \sled{}-based models can use any future pretrained LM out-of-the-box without requiring additional pretraining to further improve performance.


Due to its simplicity, \sled{} can also be used as a diagnostic tool for analyzing long-range benchmarks. 
We analyze the seven datasets in \SCROLLS{} through the lens of \sled{}
and show which datasets require the input to be contextualized with remote tokens. Specifically, we find that in QA and NLI tasks, relatively local contextualization is sufficient for high performance.

While SLED is similar to FiD from a technical standpoint, past usage of FiD has centered around open-domain question answering \cite{Izacard2021LeveragingPR}, where unrelated passages are naturally encoded independently. Here, we test fusion-in-decoder on long documents, where local encoding of chunks is a modeling assumption that needs testing. 
In recent work, \newcite{vig-etal-2022-exploring} proposed a similar architecture to tackle long inputs from QMSum \cite{zhong2021qmsum}, but did not systematically analyze it. 
We standardize this methodology for the first time, and extensively analyze the effectiveness of FiD for encoding long documents across multiple tasks.

To summarize, our main contributions are:
\begin{enumerate}[itemsep=0pt,topsep=0pt]
    \item We present \sled{}, a simple and effective approach for processing long texts that leverages off-the-shelf encoder-decoder LMs based on fusion-in-decoder.
    \item We demonstrate \sled{}'s efficacy in both controlled experiments, as well as on the \SCROLLS{} benchmark, which leads to competitive results compared to specialized models that include up to 50x more parameters.
    \item We use \sled{} as a diagnostic tool for analyzing the long-range properties of datasets in the \SCROLLS{} benchmark.
    \item We provide an open-source implementation of \sled{},\footnote{
        \iftaclpubformat
            \url{https://github.com/Mivg/SLED}
        \else
            \url{hidden-for-anonymous-submission}
        \fi
    } seamlessly integrated into the  Transformers library \cite{wolf-etal-2020-transformers}. 
\end{enumerate}

\section{Background}
\label{sec:background}

Recent advances in natural language processing have been by and large fueled by the transformer architecture \cite{Vaswani2017AttentionIA}. A core component of the transformer is the self-attention layer where every input token ``attends'' to every other token to produce its contextualized representation. This results in quadratic time and space dependency w.r.t. the length of the input, limiting the ability of transformers to process long sequences.

This long-text limitation has sparked ample interest in developing efficient transformer variants.
One prominent family of methods is based on \emph{sparse attention}, where each token attends to a constant number of other tokens, overcoming the quadratic dependency. Tokens typically attend either to their \emph{local} surrounding \cite{Zaheer2020BigBT,beltagy2020longformer,ainslie-etal-2020-etc,gupta2020gmat} or to tokens that are semantically similar \cite{Kitaev2020Reformer,roy-etal-2021-efficient}.
Moreover, a constant number of \emph{global tokens} that attend to and are attended by all input tokens are often added to each attention sub-layer.
Recent analyses \cite{xiong-etal-2022-simple} have shown that sparse transformers with local attention are competitive with other variants on multiple language understanding tasks.

Our method, SLED, falls into the family of local attention variants. However, unlike prior work, SLED re-uses and extends existing short-range encoder-decoder models, and does not require specialized pretraining or dedicated CUDA implementations. 

In most local attention variants, e.g., \led{} \cite{beltagy2020longformer}, attention is local \emph{per-layer}, but the receptive field of tokens grows \emph{across layers}. In \sled{}, which we describe next, tokens have access to the same number of tokens, independent of a layer's depth, which enables better parallelization.
For a survey on the families of efficient transformers, see \newcite{tay2020efficient}. For an in-depth comparison of \sled{} and \led{}, we refer to Appendix~\ref{app:led}.

\section{Method}
\label{sec:method}

In this work, we propose a simple approach for avoiding transformer's quadratic complexity, motivated by the \localhypo: 
\begin{displayquote}
In an encoder-decoder architecture, the encoder can effectively contextualize input tokens with local context only, leaving long-range dependencies to be handled by the decoder.
\end{displayquote}

\sled{} relies on said modeling assumption to encode shorter chunks independently and perform fusion of information in the decoder \cite{Izacard2021LeveragingPR}.
We now describe the SLED model in detail.

\paragraph{Input} \sled{} uses a pretrained encoder-decoder model $M$ as a backbone. \sled{} receives a tokenized document of length $n$ (blue squares in Fig.~\ref{fig:sled}), and an optional short tokenized prefix of length $m \ll n$, typically representing a question about the document, an instruction to perform some generation task, or a hypothesis (orange squares in Fig.~\ref{fig:sled}). Unlike static task-specific prefixes (e.g., ``summarize''), SLED supports also sample-specific prefixes that are part of the input (e.g., the question in QA datasets).

\paragraph{Steps} \sled{} follows the following steps:
\begin{enumerate}[label=(\alph*)]
\item Document tokens are split into $C$ chunks of length $c$ (In Fig.~\ref{fig:sled}, $c=4$). 
The \emph{middle} $(1-\rho) \times c$ tokens in each chunk are contextualized from both the left and right by $P:=\frac{\rho\times c}{2}$ tokens, where 
$\rho\in[0,0.5]$ ($\rho=0.5$ in Fig.~\ref{fig:sled}). We call these middle tokens \emph{the effective chunk}, since they will constitute the output of the encoder, and term the tokens on each side by \emph{context padding}.
\item Each chunk is prepended by (optional) prefix tokens (Fig.~\ref{fig:sled}(b)).
\item Each chunk is encoded independently, using the backbone encoder $M_\text{enc}$ (see  Fig.~\ref{fig:sled}(c)).
\item To create a contextualized representation for each token, we keep from each chunk only the tokens from the \emph{effective chunk}, and concatenate them (Fig.~\ref{fig:sled}(d)). 

\item To give the decoder access to prefix tokens, we encode the prefix tokens with $M_\text{enc}$, and prepend the result to the contextualized representation (leftmost chunk in Fig.~\ref{fig:sled}(a)-(d)).
\item Finally, we generate the output with the backbone decoder, $M_\text{dec}$, which uses standard cross-attention over the $m+n$ encoded tokens (Fig.~\ref{fig:sled}(e)).
\end{enumerate}

\sled{} requires handling a few edge cases, namely, dealing with the first and last chunk that do not have bidirectional context. We refer to Appendix~\ref{app:sled} for these details.

\paragraph{\sled{}'s Complexity} \label{subsec:complxity} \sled{} divides an input of length $n$ to $C$ chunks of size $c$. Since $\rho \in [0,0.5]$, it follows that $C \in \left [\frac{n}{c}, \frac{2n}{c} \right]$.
While the complexity of encoding each chunk is quadratic in $c$ due to self-attention, $c \ll n$ is constant and thus the memory and compute dependency is linear in $n$.\footnote{We assume the prefix length ($m$) is negligible
and thus its effect on asymptotic complexity is negligible.} In particular, the complexity to encode the input with a model of $l$ attention layers is:
\[
\mathcal{O}\left(l\times c^2\times\frac{2n}{c}\right)=\mathcal{O}\left(l \times c \times n\right).
\] 
Decoding is done as proposed by \citet{Vaswani2017AttentionIA}, thus requiring $\mathcal{O}(nk+k^2)$ memory. Assuming a constant output sequence length $k \ll n$, this remains linear in $n$.

\section{Efficacy of Fusion in Decoder}
\label{sec:fid}

As mentioned (\S\ref{sec:method}),
\sled{} relies on the assumption that chunks can be encoded independently and fusion across them can be delegated to the decoder (\localhypo{}).
This is similar to the Fusion-in-Decoder approach, introduced by \newcite{Izacard2021LeveragingPR} for open-domain question answering (ODQA). However, there, the encoder-decoder receives a set of independent passages and needs to generate an answer that can typically be extracted from a single passage. Here, we extend the scope of FiD by applying it over a single, long, and coherent input that potentially requires global contextualization.

To demonstrate the viability of FiD for long text language tasks, we design two controlled experiments that quantify the extent to which FiD can perform two operations at the heart of long-text processing.
First, can FiD find a ``needle-in-a-haystack'', i.e., locate a piece of short information embedded in long text, disregarding irrelevant information. Second, can FiD ``piece the puzzle'' and fuse two pieces of information that are encoded independently when generating an output. 

\subsection{Needle in a haystack}
\label{subsec:squad}
To check if \sled{} can ignore irrelevant text and locate a single piece of information, we cast SQuAD 1.1 \cite{Rajpurkar2016SQuAD1Q} as a sequence-to-sequence task with long input.
SQuAD is a question answering dataset, where given a question-paragraph pair the goal is to generate the answer (which lies within the paragraph). For each question-paragraph pair, we randomly sample 9 other paragraphs from the the dataset and concatenate them to form a long document.\footnote{We only consider paragraphs that are not within the gold document and do not contain the gold answer.} We then finetune and evaluate our models in two settings: a) \emph{Ordered Distractors}: the gold paragraph is the first one, and all other distractors are concatenated after it. b) \emph{Shuffled Distractors}: we randomly shuffle the order of all paragraphs so the answer can be anywhere in the input document. Since this is a QA task, the prefix  is the question.

We use \bartbase{} \cite{Lewis2020BARTDS} as our backbone model, $M$, throughout \S\ref{sec:fid},
and compare \sled{} to an \emph{oracle} \bartbase{} that is given the gold paragraph only with no distractor paragraphs. this is an oracle setup since \bartbase{} can take 1,024 tokens as input and all gold paragraphs are shorter.
If \sled{} can match the oracle performance, we can infer that indeed the decoder can find a needle in a haystack. In addition, we compare \sled{} to \bartbase{} which is given only the first 1K tokens and to \led{} \cite{beltagy2020longformer}, which uses local sparse attention, similar to \sled{} (\led{} has the same backbone \bartbase{}). However, as explained in \S\ref{sec:background}, the receptive field of \led{} layers linearly grows with the number of layers, and thus information can be fused in the encoder, unlike \sled{} where cross-chunk fusion must be delegated to the decoder. Last, for QA tasks, \led{} defines the question tokens as \emph{global tokens}, and as an additional sanity test we evaluate \ledlocal{}, i.e., a local LED model where no global tokens are used. For both \led{} and \sled{} we use a chunk size $c=256$.

\begin{figure}[t]
\begin{center}

\centerline{
\includegraphics[width=\linewidth,trim={0cm 0cm 0cm 0cm},clip]{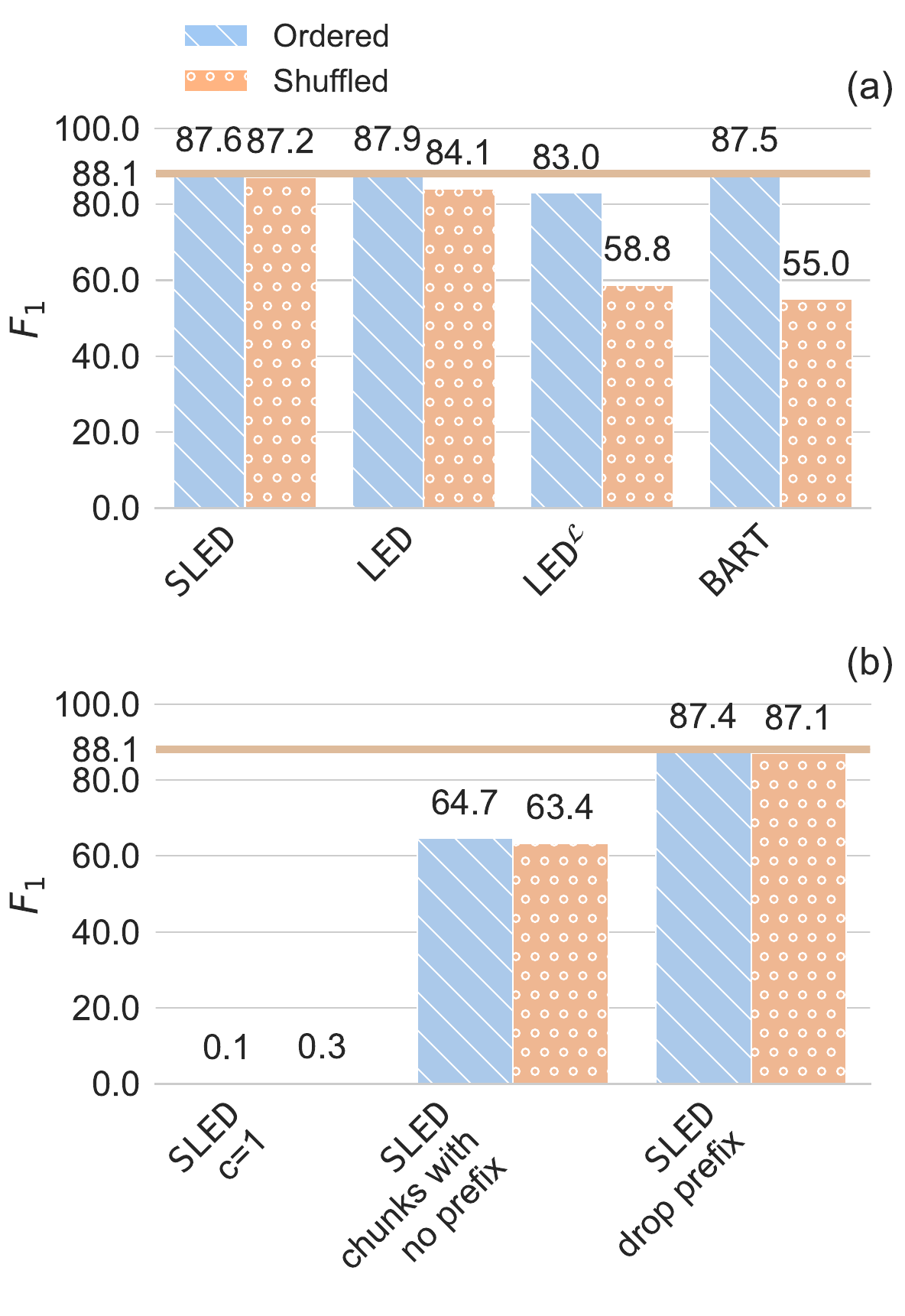}

}
\setlength{\abovecaptionskip}{-10pt}
\setlength{\belowcaptionskip}{-20pt}
\caption{$F_1$ results on our modified SQuAD 1.1's \cite{Rajpurkar2016SQuAD1Q} development set evaluation: 
(a) the horizontal line gives the performance of an oracle \bartbase{} given the gold paragraph only. \sled{} matches oracle performance in both the ordered and shuffled setting (see text). LED slightly underperforms \sled{} in the shuffled setup. Both BART (given only the first 1K tokens) and LED with no global tokens (LED$^\mathcal{L}$) performs poorly in the shuffled setup.
(b) Ablations on \sled{}'s architecture, see \S\ref{subsec:fid-ablations} for details.
}

\label{fig:squad}
\end{center}
\end{figure}

\paragraph{Results} 
Fig~\ref{fig:squad}(a) shows the results of our evaluation on the development set. 
\sled{} almost matches the performance of an oracle \bartbase{} that is not given any distractor paragraphs, reaching an F$_1$ score of 87.6 compared to the oracle F$_1$ of 88.1 (horizontal line in the figure).
LED also achieves high performance (but lower than \sled{} in the shuffled setup), showing both models learn to ignore distracting information and find a needle in a haystack.
As expected, both \ledlocal{} and BART suffers a significant drop in performance when the passages are shuffled, as the gold paragraph is not contextualized with the question.

\subsection{Piecing a puzzle} 
\label{subsec:hotpot}
\begin{figure}[t]
\begin{center}

\centerline{
\includegraphics[width=\linewidth]{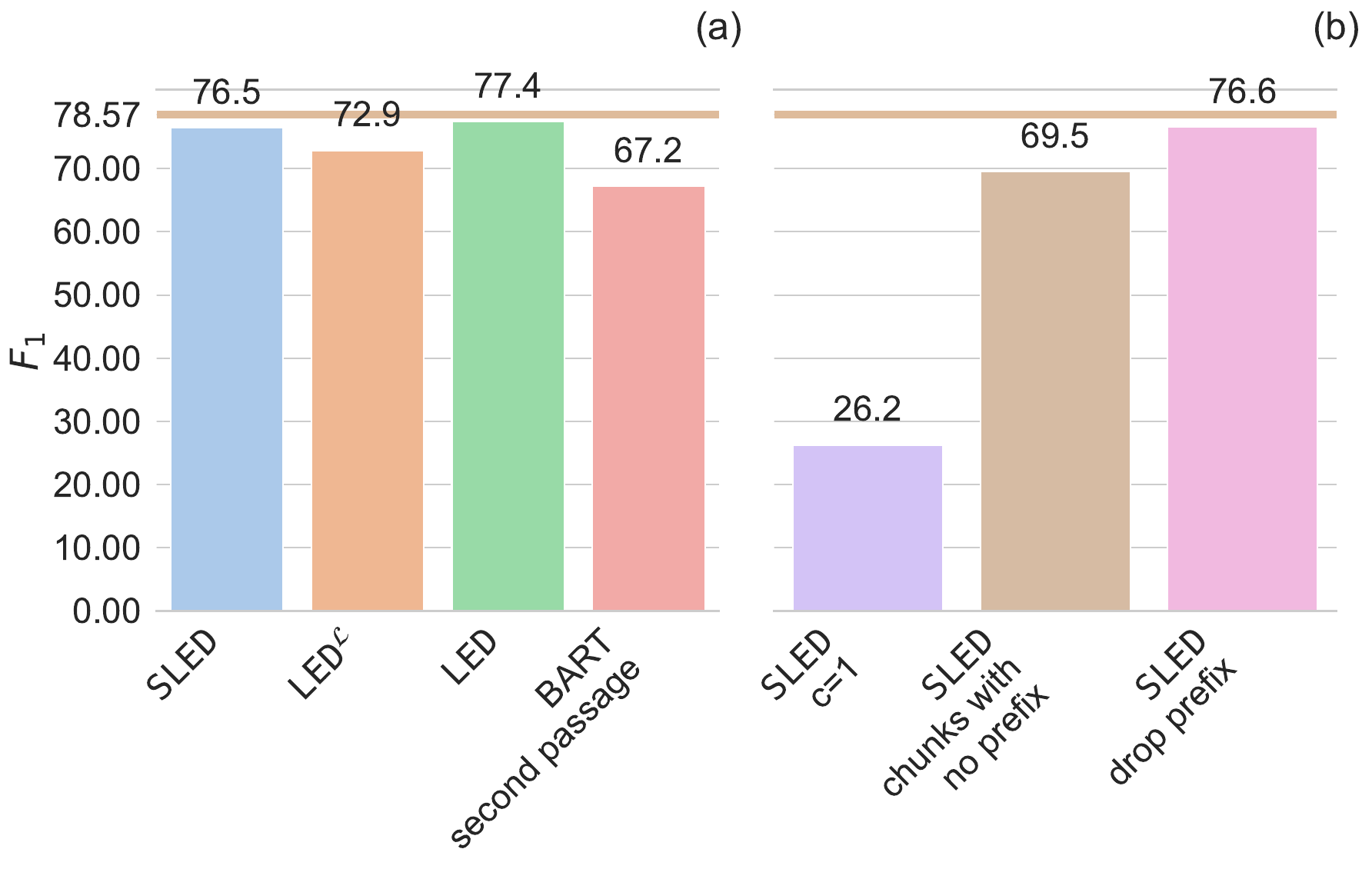}
}
\setlength{\abovecaptionskip}{-5pt}
\setlength{\belowcaptionskip}{-20pt}
\caption{$F_1$ results on our HotpotQA's development set \cite{Yang2018HotpotQAAD}. 
(a) \sled{} reaches an F$_1$ that is close to the oracle \bartbase{} (horizontal line), outperforming a model with access to the paragraph that contains the answer (``second paragraph''). This shows that \sled{} effectively fuses information from two chunks. See text for further explanation on each model.
(b) Ablations on \sled{}'s architecture, see \S\ref{subsec:fid-ablations} for details.
}

\label{fig:hotpot}
\end{center}
\end{figure}

We now verify that \sled{} can fuse pieces of information from different chunks.
To this end, we modify HotpotQA \cite{Yang2018HotpotQAAD}, a  multi-hop question answering dataset, in which every question relies on two pieces of information (located in different paragraphs). While in the original setting, each input in HotpotQA has two gold paragraphs and 8 distractor paragraphs, we include only the two gold paragraphs in our experiments. To ensure \sled{} and \led{} encode the relevant two pieces of information in \emph{separate} chunks, we set the chunk size to $c=128$.

Similar to \S\ref{subsec:squad}, we compare \sled{} to an oracle \bartbase{} with full attention over 1,024 tokens,\footnote{All examples have $\leq$1,024 tokens, including the prefix.} to LED, and to \ledlocal{}. Finally, past work has shown that many examples in HotpotQA can be answered with access to the ``second'' gold paragraph only, which contains the answer \cite{Jiang2019AvoidingRS}. Thus, we also evaluate a BART model that is given the second passage only.

\paragraph{Results} Fig.~\ref{fig:hotpot}(a) shows that indeed, \sled{}'s decoder can effectively fuse information from two separately encoded chunks, reaching an F$_1$ of 76.5, slightly lower than the oracle F$_1$ of 78.6. Notably, \sled{} substantially outperforms a BART model with access to the entire second paragraph, showing that information is fused by the decoder.
LED slightly outperforms \sled{}, but when denied access to global tokens (\ledlocal{}) its performance drops sharply.
This shows that the large receptive field of deep \led{} layers does not suffice for information fusion and interaction between the question and text is crucial for the decoder.

To summarize, our two controlled experiments show that \sled{} can perform the operations of retrieving and fusing information, which are fundamental for long text language tasks.

\subsection{Ablations of design choices}
\label{subsec:fid-ablations}
We leverage our controlled experimental setup to further investigate the components of \sled{}.

\paragraph{Efficacy of the encoder} 
While \S\ref{subsec:hotpot} shows that \sled{} can fuse separate pieces of information in the decoder, it is not clear to what extent local contextualization is necessary.
To check whether it is possible for all fusion to occur in the decoder, we finetune 
\sled{} with a chunk size of $c=1$, such that input tokens do not observe any context in the encoder. As can be seen in the leftmost bar(s) in Fig.~\ref{fig:squad}(b) and Fig.~\ref{fig:hotpot}(b), removing local contextualization results in poor performance, illustrating the importance of local contextualization.

\paragraph{Contextualizing chunks with a prefix}
As explained, \sled{} does not use global tokens, but instead contextualizes each chunk with a prepended prefix. 
To verify its necessity, we finetune a \sled{} model that treats the prefix as another chunk and does not prepend it to document chunks.\footnote{We add masked padding after the prefix to ensure chunking of the document remains identical.} The second bar(s) in Fig.~\ref{fig:squad}(b) and Fig.~\ref{fig:hotpot}(b) shows a significant drop in performance for all settings, suggesting the prefix is needed during encoding.

As expected, there is practically no difference between the \emph{Ordered} and \emph{Shuffled} settings in Fig.~\ref{fig:squad}(b). In contrast, LED$^{\mathcal{L}}$ which is similar in concept (due to the lack of global tokens) shows a significant drop when paragraphs are shuffled. This shows the possible effectiveness of the increased receptive field in LED, but only when the gold paragraph is relatively close to the prefix.

\paragraph{Encoding the prefix}
After showing the prefix is crucial for the encoder, we ask whether the decoder needs direct access to the prefix or whether relevant information from the prefix can be infused into the chunk representations. To test that, we finetune \sled{} as usual, but remove the prefix tokens from the final representation given to the decoder. The rightmost bar(s) in Fig.~\ref{fig:squad}(b) and Fig.~\ref{fig:hotpot}(b) shows that providing the decoder with prefix representations makes little difference if any at all, suggesting that indeed the encoder can infuse the important information from the prefix into the encoded document tokens.

\section{Experiments}
\label{sec:experiemnts}

We evaluate \sled{} on \SCROLLS{} \cite{Shaham2022SCROLLSSC}, a recently-proposed benchmark for evaluating long text understanding.
\SCROLLS{} contains seven datasets that span three different language understanding tasks:
\begin{enumerate}
    \item Summariazation: \emph{GovReport} \cite{huang2021govreport} is a summarization task over reports from the Congressional Research Service; \emph{SummScreenFD} \cite{chen2021summscreen} is a summarization dataset over TV scripts; \emph{QMSum} \cite{zhong2021qmsum} is a query-based summarization dataset over meeting transcripts from various domains. While GovReport and SummScreenFD do not contain a prefix, for QMSum we consider the query as the prefix.
    \item Question answering (QA): \emph{Qasper} \cite{dasigi2021qasper} is a QA benchmark that contains questions over NLP papers; \emph{NarrativeQA} \cite{kocisky2018narrativeqa}
    contains questions over entire books and movie scripts; \emph{QuALITY} \cite{pang2021quality} is a multiple-choice QA dataset over books and articles.
    For all QA datasets, we set the question as the prefix. For QuALITY, we consider the four answer options part of the question.
    \item Natural language inference: \emph{ContractNLI} \cite{koreeda-manning-2021-contractnli-dataset} contains short legal hypotheses (set as the prefix) and legal documents as the premise. Models are tasked to predict whether the premise entails, contradicts or is neutral w.r.t. to the hypothesis.
\end{enumerate}

For each task, we use the official evaluation metrics defined in \SCROLLS{}, which are based on the metrics from the original datasets.

\subsection{Settings}

We evaluate \sled{} with both BART \cite{Lewis2020BARTDS} and T5 \cite{Raffel2020ExploringTL} as backbone models.
For each backbone model, we compare performance with \sled{}, which can consume long sequences, vs. the backbone models alone that are fed with the first 1,024 tokens. For comparison, we also finetune \ledbase{}. In all \sled{} and \led{} experiments, we use a maximal sequence length of 16K tokens and chunk size of 256 to allow for a fair evaluation.

For each model-dataset pair, we run hyperparameter tuning (detailed in Appendix~\ref{app:experimental-details}) based on the development set.
Additionally, we submit generated predictions over the test set to \SCROLLS{} leaderboard,\footnote{\label{fn:leaderboard}\url{https://www.scrolls-benchmark.com/leaderboard}} and compare to the reported performance of other models at the time of submission.
\newcommand\leaderboardfootnote{\value{footnote}}

\subsection{Results}
\label{subsec:results}
\begin{table*}[th]
\footnotesize
\addtolength{\tabcolsep}{-2pt}  
\centering
\resizebox{\linewidth}{!}{\begin{tabular}{@{}l@{}rccccccccc@{}}
\toprule
\multirow{2}{*}{\textbf{Model}} & \multirow{2}{*}{\textbf{~(Chunk/Input)}} &
\multirow{2}{*}{\textbf{\#Params}} & \multirow{2}{*}{\textbf{Avg}} & \textbf{GovRep} & \textbf{SumScr} &  \textbf{QMSum} & \textbf{Qspr}  & \textbf{Nrtv}  & \textbf{QALT}  & \textbf{CNLI}  \\
&  & & & ROUGE-1/2/L  &  ROUGE-1/2/L  &  ROUGE-1/2/L  & F1  & F1 & EM-T/H &  EM \\
\midrule

\multicolumn{11}{c}{\textbf{Development Scores}} \\
\midrule

LED$_{\mathtt{base}}$ & $(256/16K)$ & 162M & - & \textbf{57.3/27.9/30.0} & 30.7/6.3/17.9 & 32.5/9.0/21.1 & 30.4 & 20.2 & 30.9 & 82.3 \\
$\mathtt{T5}_{\mathtt{base}}$ & $(1K/1K)$ & 220M & - & 32.8/11.7/20.2 & 22.2/3.7/15.3 &  26.1/6.6/19.8 & 13.2 & 14.9 & 35.1 & 76.8 \\
$\mathtt{T5}_{\mathtt{base}}$-$\mathtt{SLED}$ & $(256/16K)$ & 220M & - &            47.0/20.2/25.2 & 25.3/5.0/16.6 &  29.9/8.7/21.4 & 38.2 & 18.2 & 34.6 & 82.4 \\
$\mathtt{BART}_{\mathtt{base}}$ & $(1K/1K)$ & 139M & - & 47.7/18.5/22.3 & 30.1/7.0/18.3 &  32.2/9.3/21.1 & 23.3 & 15.9 & 33.8 & 78.4 \\
$\mathtt{BART}_{\mathtt{base}}$-$\mathtt{SLED}$ & $(256/16K)$ & 139M & - & 55.7/24.8/25.8 & 33.6/8.5/19.2 & 34.4/11.5/22.7 & 35.8 & 21.3 & 33.7 & \textbf{85.3} \\
$\mathtt{BART}_{\mathtt{large}}$ & $(1K/1K)$ & 406M & - & 50.6/19.8/23.5 & 32.1/7.4/18.7 &  33.3/9.4/21.6 & 24.5 & 17.9 & 36.1 & 79.3 \\
$\mathtt{BART}_{\mathtt{large}}$-$\mathtt{SLED}$ & $(256/16K)$ & 406M & - & \textbf{57.4}/26.3/27.5 & \textbf{35.3/8.8/19.5} & \textbf{36.3/12.2/23.3} & \textbf{42.5} & \textbf{23.6} & \textbf{37.2} & \textbf{85.3} \\

\midrule
\multicolumn{11}{c}{\textbf{Test Scores}} \\
\midrule

LED$_{\mathtt{base}}$ & $(256/16K)$ & 162M & 33.6 & 56.8/\textbf{27.3/29.2} & 30.0/6.0/17.5 & 31.3/8.6/20.5 & 34.8 & 21.0 & 28.5/28.3 & 82.9 \\
$\mathtt{T5}_{\mathtt{base}}$ & $(1K/1K)$ & 220M & 26.3 & 33.2/12.1/20.4 & 21.4/3.6/15.0 &  24.2/5.9/18.6 & 16.3 & 15.0 & 31.9/28.6 & 76.3 \\
$\mathtt{T5}_{\mathtt{base}}$-$\mathtt{SLED}$ & $(256/16K)$ & 220M & 33.3 & 46.6/20.1/25.1 & 24.5/4.6/16.5 &  28.4/8.7/20.5 & 43.0 & 18.9 & 31.2/29.4 & 81.4 \\
$\mathtt{BART}_{\mathtt{base}}$ & $(1K/1K)$ & 139M & 30.6 & 48.0/19.1/22.7 & 30.1/6.6/18.1 & 31.2/9.1/20.3 & 27.6 & 16.0 & 32.5/31.6 & 77.1 \\
$\mathtt{BART}_{\mathtt{base}}$-$\mathtt{SLED}$ & $(256/16K)$ & 139M & 35.4 & 54.7/24.4/25.4 & 32.7/7.9/19.1 &  33.8/\textbf{11.7/22.6} & 41.1 & 21.5 & 29.7/30.4 & 85.6 \\
$\mathtt{BART}_{\mathtt{large}}$ & $(1K/1K)$ & 406M & 32.1 & 50.7/20.1/23.5 & 31.6/6.8/18.5 & 32.0/9.1/20.8 & 29.2 & 18.3 & \textbf{34.8}/33.9 & 79.7 \\
$\mathtt{BART}_{\mathtt{large}}$-$\mathtt{SLED}$ & $(256/16K)$ & 406M & \textbf{38.0} & \textbf{57.5}/26.3/27.4 & \textbf{35.2/8.7/19.4} & \textbf{34.2}/11.0/22.0 & \textbf{46.9} & \textbf{24.1} & \textbf{34.8/34.8} & \textbf{87.3}\\[0.1cm]
\hdashline\\[-0.3cm]
LED$_{\mathtt{base}}^{\mathtt{SCROLLS}\dagger}$ & $(1K/16K)$ & 162M & 29.2 & \textbf{56.2}/26.6/28.8 & 24.2/4.5/15.4 & 25.1/6.7/18.8 & 26.6 & 18.5 & 25.8/25.4 & 71.5 \\
LongT5$_{\mathtt{base}}^\dagger$ & $(255/16K)$ & 220M & 38.2 & 53.5/27.3/29.3 & 34.8/\textbf{9.6}/21.1 & 33.9/11.0/22.8 & 46.6 & 23.0 & 37.9/36.6 & 85.6 \\
LongT5$_{\mathtt{large}}^\dagger$ & $(255/16K)$ & 770M & 40.5 & 54.2/27.8/29.8 & 35.6/9.2/\textbf{21.2} & 35.1/\textbf{12.0}/23.3 & 52.3 & 27.2 & 40.6/38.6 & 87.3 \\
LongT5$_{\mathtt{XL}}^\dagger$ & $(255/16K)$ & 3B & \textbf{41.9} & 54.7/\textbf{28.2/30.2} & \textbf{35.8/9.6}/21.1 & \textbf{34.9}/11.8/\textbf{23.5} & \textbf{53.1} & \textbf{29.3} & \textbf{46.0/42.1} & 88.2 \\
UL2$^\dagger$ & $(2K/2K)$ & 20B & 37.9 & 53.6/26.1/28.8 & 32.9/7.8/19.4 & 31.1/8.5/20.4 & 37.6 & 24.2 & 45.8/40.7 & \textbf{88.7}\\

\bottomrule
\end{tabular}}
\caption{Main results on the \SCROLLS{} benchmark. Chunk/Input refers to the chunk size used ($c$) and to the maximal input length ($n$). Avg is the average \SCROLLS{} score as described in \newcite{Shaham2022SCROLLSSC}. Development scores for QuALITY are only for the full set (T). $\dagger$ indicates reported results from \SCROLLS{} public leaderboard.\footnotemark[\leaderboardfootnote{}]
\ledbaseglobal{} scores were reported by \newcite{Shaham2022SCROLLSSC} and are lower than 
our LED$_{\mathtt{base}}$ implementation, presumably since our implementation uses all question tokens for global attention rather than just the first one. The results for LongT5  and UL2 were submitted to the SCROLLS leaderboard by their authors.}

\label{tab:main_results}
\end{table*}

Tab.~\ref{tab:main_results} reports results over \SCROLLS{} development and test sets.
Taking short-range pretrained LMs like BART and T5 and casting them into \sled{}'s framework allows them to process long documents effectively, improving the average \SCROLLS{} score by 4.8-7 points.
Examining \bartbase{}-\sled{}, we see a large improvement compared to \ledbase{} (33.6$\rightarrow$35.4), and competitive performance on multiple tasks compared to LongT5$_{\mathtt{base}}$ and UL2.
Moreover, adding \sled{} to \bartlarge{}
results in a high-performing model
with results that are comparable to LongT5$_{\mathtt{base}}$ and outperforming UL2, despite UL2's large parameter count (50x larger), and with no need for expensive pretraining geared towards long-range tasks. \bartlarge{}-\sled{}'s performance is moderately lower than the larger LongT5 models.

Barring QuALITY, \sled{} significantly improves performance across all tasks compared to the corresponding backbone models.
All summarization datasets (GovReport, SummScreenFD and QMSum) show impressive gains of up to 35\% compared to their baseline scores, across all metrics (Rouge-1/Rouge-2/Rouge-L \cite{Lin2004ROUGEAP}) and for all three backbone models. 
Similarly, on ContractNLI \cite{koreeda-manning-2021-contractnli-dataset} we see large relative improvements. As the performance of the baseline models was already high, this boost in performance is even more significant.
Finally, the QA datasets Qasper and NarrativeQA show the largest gains, improving by an average of 60\%.

\paragraph{QuALITY} In stark contrast to other datasets lies the multi-choice QA dataset QuALITY \cite{pang2021quality}. While the performance of \bartlarge{}-\sled{} is above chance, it barely improves the performance of its backbone model (\bartlarge{}), which observes only the first 1K tokens, with a similar trend in other backbone models. Analyzing test scores in Tab.~\ref{tab:main_results}, we see  that
increasing model size consistently improves performance (up to 46\% exact match), but increasing input length has a negligible effect. 
Since reported human accuracy on QuALITY is high (93.5\%), this hints that QuALITY might require commonsense reasoning and knowledge that are absent from models with a lower parameter count.

\paragraph{Summary}
We have shown that taking off-the-shelf pretrained LMs and embedding them into \sled{} leads to competitive performance on \SCROLLS{}. Importantly, any future pretrained LM can be easily plugged into \sled{}, without the need for an expensive pretraining step.

\subsection{Datasets analysis}
\sled{}'s simplicity and modularity allow it to be used as a useful tool for dataset analyses. Specifically, we can vary the chunk size, $c$, and the number of tokens, $n$, across datasets to analyze a) how local are individual pieces of relevant information, and b) 
how far into the document they are located.

\begin{figure}[t]
\begin{center}

\centerline{
\includegraphics[width=\linewidth]{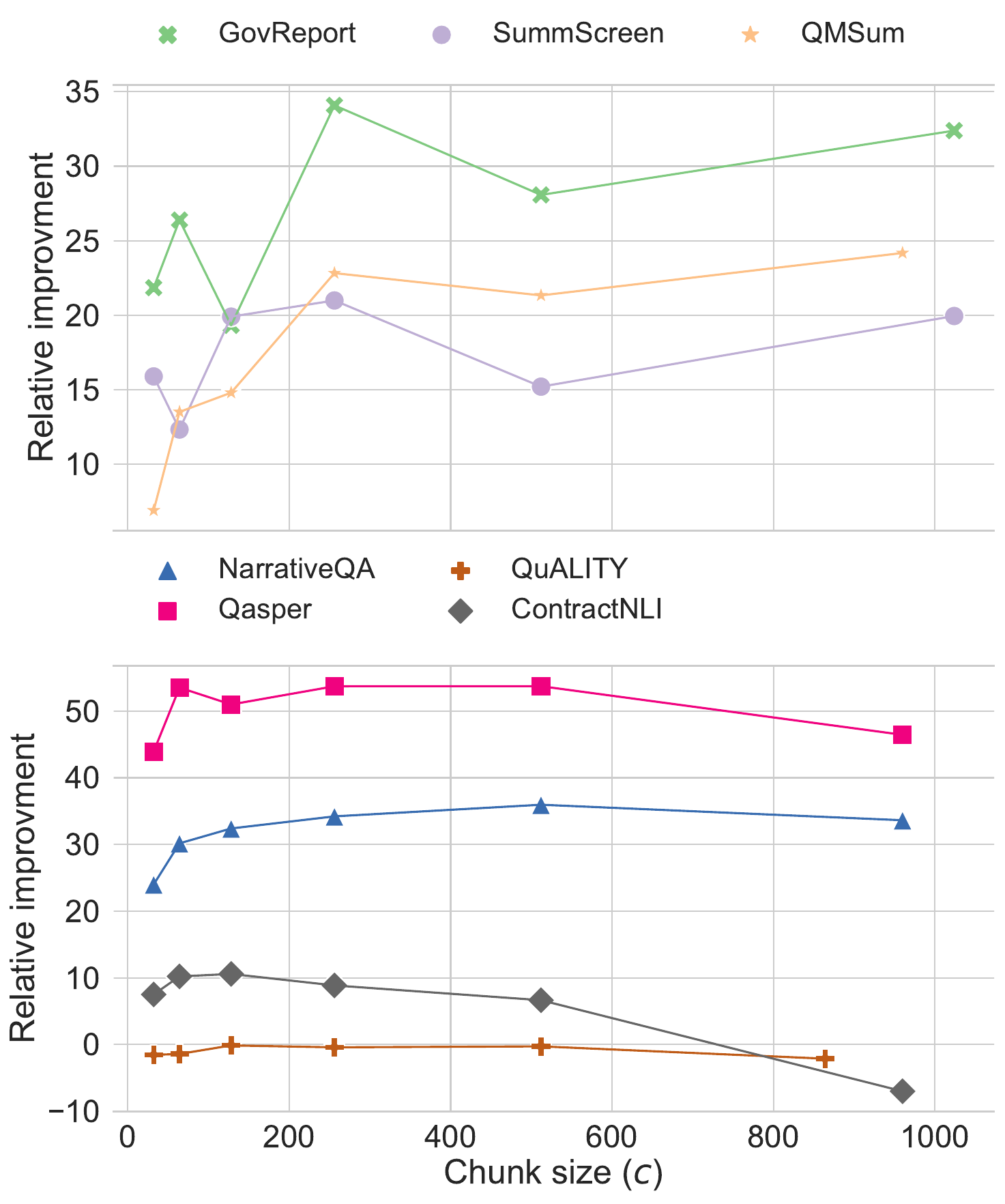}
}
\setlength{\belowcaptionskip}{-20pt}
\caption{\bartbase{}-\sled{} relative improvement compared to \bartbase{} results, when varying the \sled{}'s chunk size (i.e. $c$), fixing the maximum input length to 16K. \textbf{Top}: Summarization datasets. The y-axis measures relative improvement of Rouge-2. \textbf{Bottom}: QA and NLI datasets. The y-axis measures relative improvement of exact match for QuALITY and ContractNLI and F$_1$ for NarrativeQA and Qasper.
}
 
\label{fig:ab-chunk}
\end{center}
\end{figure}

\label{subsec:properties}
\paragraph{Locality of information}
\sled{} relies on an assumption that information can be contextualized locally at encoding time. To analyze locality, we vary the chunk size, $c$, which defines the attention window, and measure
the effect on \SCROLLS{} datasets with input length 16K. 
Fig.~\ref{fig:ab-chunk} shows the results of this experiment, where the y-axis shows the relative improvement compared to \bartbase{} on a target metric as a function of the chunk size $c$ for all datasets.
We observe that in all datasets the best performing chunk size is relatively small (up to 256), and further increasing $c$ even hurts the performance in some cases. However, the summarization datasets show a much larger gain in performance when increasing $c$ up to that threshold.
This coincides with a common hypothesis that QA and NLI require relatively local context, and thus increasing $c$ can add noise and hurt optimization, while summarization may require a more high-level view of information.

\begin{figure}[t]
\begin{center}

\centerline{
\includegraphics[width=\linewidth]{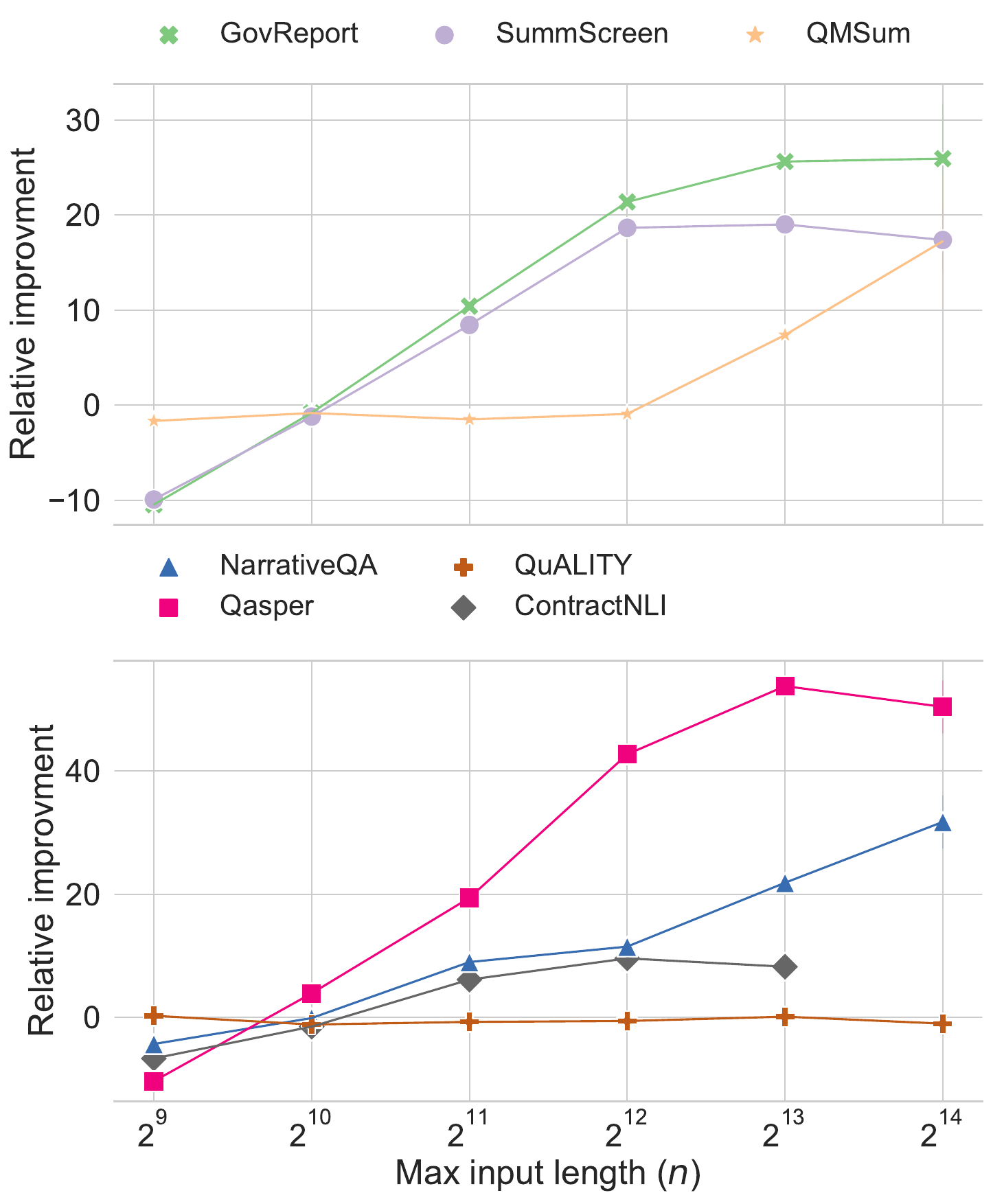}
}
\setlength{\belowcaptionskip}{-20pt}
\caption{\bartbase{}-\sled{} relative improvement compared to \bartbase{} results, when varying the input length fed to \sled{}, fixing $c=256$. \textbf{Top}: Summarization datasets compared w.r.t. Rouge-2. \textbf{Bottom}: QA and NLI datasets. Relative improvment is measured w.r.t. exact match for QuALITY and ContractNLI and F$_1$ for NarrativeQA and Qasper.}

\label{fig:ab-length}
\end{center}
\end{figure}

\paragraph{Distance from start of document}
We now analyze whether indeed the entire document is required for tasks in \SCROLLS{} by varying the maximum document length, $n$.
Fig.~\ref{fig:ab-length} shows the results of this experiment, where the y-axis shows relative improvement of \bartbase{}-\sled{} compared to \bartbase{} as a function of the first $n$ tokens from the document (chunk size $c=256$). As expected, all datasets (except QuALITY) show a roughly monotonic improvement in performance with $n$. 
This shows that
(a) \sled{} is able to effectively use all of the information in a long sequence (up to 16K tokens),\footnote{For ContractNLI, the length of over 95\% of the tokenized examples is less than 8K.} and that (b) observing the entire inputs from \SCROLLS{} improves performance.

\subsection{Effect of context padding}
In all experiments thus far, we used a conservative padding value $\rho=0.5$, resulting in effective chunk size of $\frac{c}{2}$ and $\frac{c}{4}$ context padding tokens on each side.
Since both memory and, more importantly, the number of forwards passes through the encoder are linear in the number of chunks, a natural question is how much padding and overlap are necessary to achieve satisfactory results. 

To explore this, we finetune \bartbase{}-\sled{} on all six datasets  where \sled{} showed gains over its baseline model (i.e., all datasets except for QuALITY), varying the value of $\rho$, and fixing $c=256$. Tab.~\ref{tab:padding} shows the results of this experiment, where we compare relative gain compared to \bartbase{} across different $\rho$ values.

As expected, decreasing the padding factor and consequently the number of chunks reduces training time. When $\rho=0.05$ training can be faster by up to 2x compared to $\rho=0.5$ as the number of chunks drops to almost half. 
Moreover, relative gain (i.e., improvement relative to the baseline) is often close to or even higher with less padding (perhaps due to better encoding or more stable optimization). Nevertheless, there is no single $\rho$ value that consistently beats the conservative choice of $\rho=0.5$. In particular, in all six datasets, setting $\rho=0.5$ results in a top-2 performance, often by a large margin and never considerably worse then the best result.
Thus, we conclude that one may improve the efficiency and performance of \sled{} by tuning the hyperparameter $\rho$ for optimal behaviour w.r.t. a specific task, and we fix $\rho=0.5$ in our experiments.

Moreover, Tab.~\ref{tab:padding} demonstrates the importance of having chunks at least partially overlapping. In all six dataset, using non-overlapping chunks ($\rho=0$) results in a drop of at least 10\% gain compared to the best setting, where in some cases this gap grows to over 50\%. This supports our hypothesis that chunking inputs with no overlap may lead to crucial loss of information.

\begin{table}[th]
\footnotesize
\addtolength{\tabcolsep}{-3pt}
\centering
\resizebox{\linewidth}{!}{\begin{tabular}{@{}lcccccc@{}}

\toprule

\multirow{2}{*}{\textbf{$\mathbf{\rho}$}} & \multicolumn{6}{c}{\textbf{Relative Gain}}\\
& \underline{GovRep} & \underline{SumScr} &  \underline{QMSum} & \underline{Qspr}  & \underline{Nrtv} & \underline{CNLI} \\

\rule{0pt}{2.5ex}50\% & \textbf{34.1\%} & \textbf{21.0\%} & \underline{22.8\%} & \underline{53.7\%} & \textbf{34.2\%} &  \underline{8.9\%} \\
25\% & \underline{28.5\%} & \underline{19.0\%} & 17.9\% & \textbf{54.7\%} & 29.4\% & \textbf{10.1\%} \\
 5\% & 18.7\% & 15.9\% & \textbf{23.5\%} & 52.0\% & \underline{31.9\%} &  7.4\% \\
 0\% & 27.1\% &  9.5\% & 11.5\% & 46.1\% & 29.2\% &  6.9\% \\

\bottomrule
\end{tabular}}
\caption{\bartbase{}-\sled{} relative improvement compared to \bartbase{} when varying the padding percentage ($\rho$).
In all cases the maximum input length is 16K and $c=256$. Relative gain is measured w.r.t. Rouge-2 for GovReport, SummScreenFD and QMSum, F$_1$ for Qasper and NarrativeQA and exact match for ContractNLI. In each column, \textbf{boldface} marks the top performing value and \underline{underline} the second-best.}
\label{tab:padding}
\end{table}

\section{Related Work}
\label{sec:related}

\paragraph{Efficient transformers}
Many efficient attention variants were proposed in recent years, to alleviate the quadratic complexity of dense attention \cite{tay2020efficient, fournier2021practical}. Among those are clustering vectors to distinct buckets, calculating attention only within each one \cite{Kitaev2020ReformerTE}, attending only to a fixed number of hidden vectors \cite{ma2021luna}, using random features to approximate the attention matrix \cite{choromanski2021rethinking,peng2021random}, and using low-rank factorizations \cite{wang2020linformer}. Despite achieving respectable performance when finetuning these models on the Long Range Arena benchmark \cite{Tay2021LongRA}, many of them were not yet proven to work well as a backbone for pretrained language models. In fact, recent work \cite{Xiong2021SimpleLA} on encoder-only models found many do \textit{not} outperform a simple local attention sliding window on downstream language tasks. We discuss such methods next.

\paragraph{Sparse attention variants} A popular and simple solution for allowing attention-based models to process long sequences is to use local attention, where each token attend to a local window around it. Longformer \cite{beltagy2020longformer}, GMAT \cite{gupta2020gmat}, and ETC \cite{ainslie-etal-2020-etc} use short windows of full attention, combined with full attention to a small number of predefined global input tokens. BigBird \cite{NEURIPS2020_c8512d14} shares the local and global features, and additionally randomly samples tokens to attend to. Finally, the recently proposed LongT5 \cite{Guo2021LongT5ET} extends T5 \cite{JMLR:v21:20-074} with local and global attention components based on ETC, relieving the need to manually specify global tokens. 
In this work, we demonstrate that a simple sliding window with off-the-shelf models without any modifications is a strong alternative for multiple generative tasks that require processing long documents.

\paragraph{Beyond transformers}
As an alternative to transformers for processing long sequences, \citet{Gu2021EfficientlyML} proposed the Structured State Space (S4) architecture showing dramatic gains over transformers on the LRA benchmark \citep{Tay2021LongRA}. 
State space models are now an active research field \cite{Gupta2022DiagonalSS,Mehta2022LongRL}, but their efficacy on long-range language understanding tasks has not been tested yet.

\paragraph{Fusion-in-Decoder} \citet{Izacard2021LeveragingPR} proposed to encode multiple independent passages separately, and concatenate the encodings prior to the decoding phase. Despite encouraging empirical evidence \citep{Amouyal2022QAMPARIA,Yavuz2022ModelingMQ}, we are the first (to our knowledge) to analyze FiD's feasibility and limitations in a controlled setting. Importantly, we test FiD on long-range tasks over a single long document, rather than a collection of independent passages.

\paragraph{Pretrained models with sliding windows}
Wrapping a BERT encoder within a sliding window was proposed by \citet{cui-hu-2021-sliding} in the context of a specialized architecture for summarization. \citet{wang-etal-2019-multi} showed that sliding BERT across text improves performance on several QA datasets. In this work, we propose a sliding window approach that can be easily plugged into any existing encoder-decoder model without additional parameters or task-specific training, and show its efficacy for long-range text understanding. 
Most similar to \sled{}, is the \textsc{SegEnc} approach proposed by \newcite{vig-etal-2022-exploring}. By dividing inputs from QMSum into overlapping chunks, encoding them separately, and then performing FiD (using two representations for every input token), the authors were able to achieve state-of-the-art results. However, \newcite{vig-etal-2022-exploring} were focused on summarization and did not perform a systematic analysis of this type of architecture.
\section{Limitations}
\label{sec:discussion}

We present \sled{} as a simple and effective method to extend the capabilities of pretrained short-text models to long-text tasks. 
Despite its impressive empirical performance on \SCROLLS{}, \sled{} suffers from two disadvantages which may limit its applicability to some long-range tasks.

\paragraph{Long output} To obtain linear complexity, \sled{} assumes the output length $k$ is constant. This is since the decoder uses quadratic self-attention over the output, on top of $\mathcal{O}(nk)$ cross-attention between the output and input. While most current long-text tasks follow this assumption, future tasks, such as academic reports or script writing, may require long text generation. This limitation is not unique to \sled{} and affects other long-range transformers including LongT5 and LED. 
Aside from finetuning, this also affects pretraining models on long inputs with self-supervised losses such as span-corruption \cite{Raffel2020ExploringTL} or denoising  \citep{Lewis2020BARTDS}, which require the decoder to process an output that is linear in the length of the input.

\paragraph{Co-reference resolution and fact retention} An assumption at the heart of \sled{} is the \localhypo{}. When the input text is long, this assumption may break if distant entity resolution or factual knowledge are required. For example, a chapter in a book may mention \emph{ ``they were walking into the room''} when knowledge of what room or who walked is located a few chapters back. In such cases, the encoder used by \sled{} will not be able to access this information, moving more responsibility to the decoder and reducing the effectiveness of the contextual encoding. Similarly, in multi-hop questions \cite{Yang2018HotpotQAAD}, attending to one part of the context is necessary in order to fully understand the question and encode a second piece of information correctly. As the encoder will not have access to the first context that leads to better question understanding, here as well more responsibility is delegated to the decoder.
\section{Conclusions}
\label{sec:conclusions}

In this work we present \sled{}, a simple approach for modeling long texts which slides a pretrained short-range encoder over a long input document and then generates an output by attending to the encoded tokens.
We show \sled{} can perform core operations that are important for long text understanding, such as finding relevant pieces of information and fusing them at decoding time, and demonstrate competitive performance on the \SCROLLS{} benchmark compared to larger models and models that employ a dedicated and expensive pretraining step. 

One of \sled{}'s most attractive features is that it can be readily used with any short-range pretrained LM. Thus, any future encoder-decoder model can be flexibly plugged into it to achieve further gains in performance on \SCROLLS{}, some of its tasks, or any other long-range task.

We open source \sled{} and hope it encourages the research community to easily extend to longer inputs and push the borders of NLU models' applicability in real-world use-cases.

\iftaclpubformat
\section*{Acknowledgements}
This research was partially supported by The Yandex Initiative for Machine Learning, the Shashua Fellowship, the Len Blavatnik and the Blavatnik Family foundation, and the European Research Council (ERC) under the European Union Horizons 2020 research and innovation programme (grant ERC DELPHI 802800). 
We would also like to thank our action editor and the anonymous reviewers for their insightful suggestions and feedback.
This work was completed in partial fulfillment for the Ph.D. degree of the first author.
\else
\fi

\appendix

\section{\sled{} implementation details}

\label{app:sled}

While \S\ref{sec:method} details \sled's method it leaves out dealing with the edge tokens for brevity. Encoding  the first and last $\frac{\rho \times c}{2}$ input tokens requires special attention, as they lack bidirectional context. 
To preserve as much commonality between chunks, all first $\frac{(2-\rho) \times c}{2}$ tokens are considered the \emph{effective chunk} tokens in the first chunk. 
To account for the final tokens, the last chunk will always start at token $t_{n-c+1}$ so it would contain exactly $c$ tokens, and its \emph{effective chunk} tokens will be defined as all tokens that were not part of any previous \emph{effective chunk}.

\section{Chunking vs. local-attention}
\label{app:led}
Both LED and SLED are long-range models built on top of the same short-text model (BART), and employ local attention. However, SLED relies on chunking, while LED uses per-layer local attention. In this section, we now discuss in more detail the relation between the two approaches.

\paragraph{Implementation} One of \sled's biggest advantages is that it is \emph{agnostic} to the backbone encoder-decoder model, and can extend any existing model without additional implementation overhead. In contrast, The attention mechanism in Longformer, and subsequently LED, was implemented by \newcite{beltagy2020longformer} with a specialized CUDA kernel that is coupled to the architecture and implementation of BART. This makes \led{} more efficient, but extending it to new architectures incurs significant engineering overhead. 
This is since LED uses a ``diagonal'' local-window attention across layers, for which a na\"ive implementation is inefficient. Conversely, SLED uses chunking, which allows to simply wrap an existing encoder-decoder model.

\paragraph{Contextualization} The most significant difference between LED and SLED from a conceptual point of view is their contextualization mechanism. While \sled{} splits the input into (overlapping) chunks and encodes each of them independently, \led{} performs local attention \emph{per-layer}. This results in an effective receptive field that grows linearly with the encoder depth, potentially allowing it to perform more ``global'' contextualization.
Our results in \S\ref{sec:fid} suggest that such global contextualization is beneficial, and a similar conclusion can be reached when observing that \ledbase{}, which uses all prefix tokens as global tokens, outperforms \ledbaseglobal{}, which uses only a single token for global contextualization.

\paragraph{Positional information} \sled's chunking mechanism means that it utilizes the positional encoding of the underlying model independently in each chunk, and is thus agnostic to the positional embedding technique used by the backbone model. Moreover, it potentially allows \sled{} to generalize to arbitrary input lengths. In contrast, \led{} utilizes BART's absolute embeddings, duplicating them 16 times to support 16K-long sequences. This limits its ability to generalize to longer inputs, and potentially induces a requirement for significant amounts of long-input samples to properly tune those new parameters \cite{Shaham2022SCROLLSSC}. This is evident in  Tab.~\ref{tab:main_results} when comparing tests scores of \ledbase{} against \bartbase{}-\sled{} and considering the number for training samples. In NarrativeQA and GovReport, which contain $\sim$71K and $\sim$19K samples respectively, \led{} is comparable to \sled{} and even slightly outperforms it on some metrics. In ContractNLI ($\sim$10K examples), it does slightly worse. In all other datasets, where the training data is small, \led{} is significantly worse than \sled{}.

\paragraph{Complexity} We analyzed the complexity analysis of \sled{}'s encoder (\S\ref{subsec:complxity}), which is $\mathcal{O}\left(l \times c \times n\right)$. A similar analysis of \led{} yields that in each layer, \led{} considers $\mathcal{O}(n)$ windows of length $c$, where in each window only the middle token attends to its local neighborhood, resulting in $\mathcal{O}\left(l \times c \times n\right)$ memory complexity as well. However, due to \sled{}'s use of overlap and full self-attention within each chunk, \sled{}'s encoding may require up 2x more memory compared to \led{} when $\rho=0.5$.
\section{Experimental details}
\label{app:experimental-details}

Our experimental setup is based on the \SCROLLS{} official repository.\footnote{\url{https://github.com/tau-nlp/scrolls}} The datasets inputs and splits remained as suggested by the authors of \SCROLLS{} as well as the suggested number of epochs per dataset.
To perform model selection, for each model-dataset pair we finetuned 9 models with LINEAR learning rate scheduling, AdamW optimizer with the default settings, and setting the learning rate to one of $\{2\mathrm{e}{-5}, 5\mathrm{e}{-5}, 1\mathrm{e}{-4}\}$ and the effective batch size to one of $\{8, 16, 32\}$. Warmup was fixed at 10\% and weight decay at 0.01.
All code, data, python environment requirements, hyperparameters and scripts required to reproduce our results will be made public upon publication.

\bibliography{tacl2021}
\bibliographystyle{acl_natbib}

\end{document}